\documentclass[twocolumn,10pt]{article}

\usepackage{amsmath,amsfonts}
\usepackage[ruled,linesnumbered]{algorithm2e}
\usepackage{algorithmic}
\usepackage[numbers]{natbib}
\usepackage{color}
\usepackage{ulem}
\usepackage{array}
\usepackage{multirow}
\usepackage{booktabs}
\usepackage{newtxtext}  
\usepackage{newtxmath}  
\usepackage[numbers]{natbib}  
\usepackage{fancyhdr}
\usepackage{graphicx}
\usepackage{mathrsfs}
\usepackage{setspace} 
\usepackage[hang,flushmargin]{footmisc}
\usepackage[a4paper, left=1.85cm, right=1.85cm, top=2cm, bottom=2cm]{geometry}
\fancypagestyle{plain}{
	\fancyhf{} 
}

\usepackage{titlesec}

\usepackage{url}
\titlespacing{\section}{0pt}{5pt}{5pt}  
\titlespacing{\subsection}{0pt}{4pt}{4pt}
\begin{document}

\title{PanoGen++: Domain-Adapted Text-Guided Panoramic Environment Generation for Vision-and-Language Navigation}

\author{
	Sen Wang\textsuperscript{*}, 
	Dongliang Zhou\textsuperscript{*}, 
	Liang Xie, 
	Chao Xu, 
	Ye Yan, and
	Erwei Yin
}
\date{}

	\twocolumn[ 
	\maketitle
	\centering
	]
	
\renewcommand{\thefootnote}{} 
\footnotetext{* The first two authors contributed equally to this work. 
	
	S. Wang is affiliated with the College of Intelligence and Computing, Tianjin University, Tianjin, China, and the Tianjin Artificial Intelligence Innovation Center, Tianjin, China. D. Zhou is with the Harbin Institute of Technology, Shenzhen, China. L. Xie, Y. Yan, and E. Yin are affiliated with the Defense Innovation Institute, Academy of Military Sciences, Beijing, China, and the Tianjin Artificial Intelligence Innovation Center, Tianjin, China. C. Xu is with the College of Intelligence and Computing, Tianjin University, Tianjin, China. Corresponding authors: Dongliang Zhou and Erwei Yin.\\ \url{https://doi.org/10.1016/j.neunet.2025.107320}}
\renewcommand{\thefootnote}{\arabic{footnote}}

\begin{abstract}
Vision-and-language navigation (VLN) tasks require agents to navigate three-dimensional environments guided by natural language instructions, offering substantial potential for diverse applications. However, the scarcity of training data impedes progress in this field. This paper introduces PanoGen++, a novel framework that addresses this limitation by generating varied and pertinent panoramic environments for VLN tasks. PanoGen++ incorporates pre-trained diffusion models with domain-specific fine-tuning, employing parameter-efficient techniques such as low-rank adaptation to minimize computational costs. We investigate two settings for environment generation: masked image inpainting and recursive image outpainting. The former maximizes novel environment creation by inpainting masked regions based on textual descriptions, while the latter facilitates agents' learning of spatial relationships within panoramas. Empirical evaluations on room-to-room (R2R), room-for-room (R4R), and cooperative vision-and-dialog navigation (CVDN) datasets reveal significant performance enhancements: a 2.44\% increase in success rate on the R2R test leaderboard, a 0.63\% improvement on the R4R validation unseen set, and a 0.75-meter enhancement in goal progress on the CVDN validation unseen set. PanoGen++ augments the diversity and relevance of training environments, resulting in improved generalization and efficacy in VLN tasks.
\end{abstract}


\maketitle

\section{Introduction}
\label{section:Introduction}
Vision-and-language navigation (VLN)~\cite{anderson2018vision,yue2024safe, wang2024vision, zhou2024navgpt, liu2024volumetric} is a pivotal task within the field of embodied intelligence~\cite{chrisley2003embodied,madl2015computational,ccatal2021robot}, concentrating on training agents to navigate through three-dimensional (3D) room environments guided by natural language instructions. Recently, it has garnered considerable scholarly attention due to its extensive commercial potential and diverse applications, such as repetitive deliveries and operations in hazardous environments. Given the scarcity of VLN training data, certain studies have prioritized data-centric research, encompassing the creation of new datasets and the data augmentation of existing ones. 
For the first aspect, room-to-room (R2R)~\cite{anderson2018vision} was the first benchmark for fine-grained navigation.  
Subsequent datasets, such as room-for-room (R4R)~\cite{jain2019stay}, extended the instructions in terms of trajectory length.
To be interaction-friendly, the cooperative vision-and-dialog navigation (CVDN) dataset~\cite{thomason2020vision} comprises human-human dialogs centered on navigation, simulating interactions between an agent and a human. 
Nonetheless, these datasets are uniformly based on the same room environments from Matterport3D (MP3D)~\cite{chang2017Matterport3D}, which encompasses only 61 interior room environments for agents' training. This limitation is due to the challenging and labor-intensive processes involved, such as capturing 360-degree panoramas for each viewpoint and manual collection and annotation. 
The restricted scale and diversity of training environments hinder agents' ability to learn effective navigation policies and generalize to unseen environments. Consequently, research has shifted towards data augmentation for VLN, which is the core focus of our study. In prior studies,
\begin{figure}[t]
  \centering
    \includegraphics[width=1.0\columnwidth]{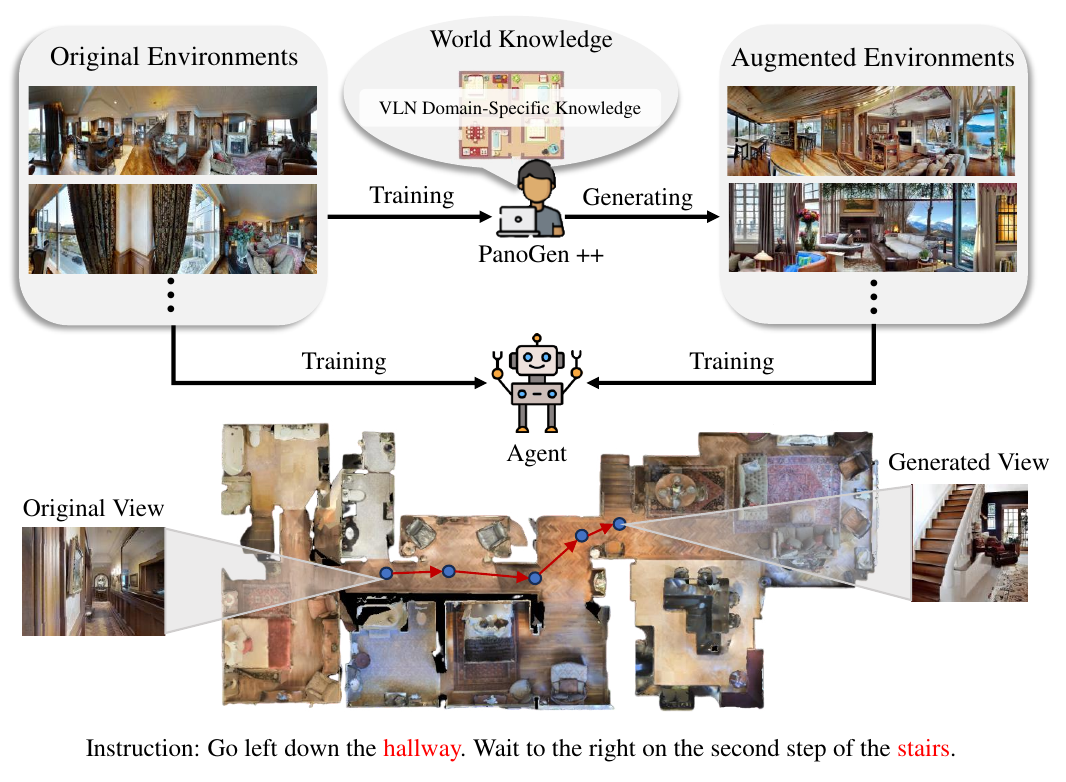}
  \caption{Domain-adapted environment generation framework (i.e., PanoGen++) tailored to MP3D for VLN agent training. In this setup, the agent receives instructions and navigates from a specified trajectory. The environment includes both original MP3D scenes and environments generated by PanoGen++, strengthening the agent's generalization to unseen settings.}
  \label{fig:intro}

\end{figure}
environmental dropout (EnvDrop)~\cite{tan2019learning}
utilizes an environmental dropout method to mimic new unseen environments, thereby enhancing the variety of observed environments. 
Random environmental mixup (REM)~\cite{liu2021vision} generates augmented data by creating cross-connected house scenes through the mixup of different environments. 
Environment editing (EnvEdit)~\cite{li2022envedit} utilizes generative adversarial networks (GANs)~\cite{goodfellow2020generative, zhou2024learning, zhou2023coutfitgan} to edit seen environments in terms of style, object appearance, and object classes.  
Despite these advancements, these methods are still confined to the indoor environments within Matterport3D.
In contrast,
Airbert~\cite{guhur2021airbert} utilizes image-caption pairs from home environments on online marketplaces to construct the pre-training task for VLN. Subsequently, AutoVLN~\cite{chen2022learning}  incorporates 900 unlabeled 3D buildings from the Habitat-Matterport 3D dataset (HM3D)~\cite{ramakrishnan2habitat}. 
To mitigate the complexity inherent in manual data construction, panoramic environment generation (PanoGen)~\cite{li2023panogen} leverages a pre-trained text-to-image (T2I) latent diffusion model~\cite{rombach2022high,hao2025lipgen}, i.e., Stable Diffusion, to generate new panoramic environments using external knowledge.  
However, the simulated VLN environments, such as Matterport3D, exhibit a distinct and consistent style, encompassing various scene types including homes, offices, and churches.
While PanoGen benefits from training on large datasets by integrating extensive world knowledge to create rich, high-quality images, it fails to address the specific requirements of VLN tasks.

To overcome these limitations, we introduce PanoGen++, a novel framework that significantly enhances PanoGen by incorporating VLN-specific domain knowledge into the generative process. Unlike PanoGen, which remains agnostic to VLN-specific characteristics, PanoGen++ achieves tailored panoramic image generation for VLN environments by integrating pre-trained diffusion models with domain-specific annotations. Initially, we employed a classical vision-language model, the bootstrapping language-image pre-training (BLIP-2) \cite{li2023blip} model, to annotate the view images from the MP3D with textual descriptions.
Subsequently, these image-text pairs were used to inject diverse knowledge of VLN into the pre-trained model of PanoGen.
However, direct fine-tuning of the entire generative model can lead to catastrophic forgetting, compromising the model's generalization capability. To address this challenge, PanoGen++ incorporates parameter-efficient fine-tuning (PEFT) techniques, which are commonly used in large language models for their computational efficiency.
Inspired by the low intrinsic dimension of over-parameterized models, we employed low-rank adaption (LoRA)~\cite{hu2022lora,kim2024hydra} technique. 
Consequently, we adapted the generation module of PanoGen to the VLN-specific model while lowering computational complexity and memory requirements by utilizing low-rank adaptation.
Our method leverages low-rank approximation to efficiently adapt the pre-trained latent diffusion model to a specific VLN environment with minimal additional weights, thus reducing computational complexity and memory requirements.
We explored two experimental settings to effectively utilize the adapted generation module for generating domain-specific panoramic environments to enhance the VLN training tasks.
The first setting, masked image inpainting, maximizes the generation of new environments by retaining only a center crop from a single panorama view and inpainting the masked areas based on corresponding text descriptions.
The second setting, recursive image outpainting, ensures the agent learns the spatial relationships between objects in the same panorama by gradually rotating the camera angle and outpainting unseen observations based on text descriptions.

Figure~\ref{fig:intro} demonstrates the capabilities of PanoGen++ in augmenting VLN environments through environment generation. It illustrates the semantic coherence between the generated observational view and the provided instruction, specifically highlighting the alignment of the ``stairs'' mentioned in the instruction with the generated environment. In particular, the agent undergoes pre-training on a combination of original Matterport3D environments and those generated by PanoGen++. During the fine-tuning phase, a subset of original views is randomly substituted with our synthetic panoramic environments.
We conducted experiments on R2R, R4R, and CVDN datasets to evaluate the performance of agents trained with PanoGen++.
The empirical results validated the effectiveness of PanoGen++ with respect to various evaluation metrics, in comparison to current state-of-the-art methods.

To sum up, the main contributions of this research are:
\begin{itemize}
	\item 
    We introduce PanoGen++, a novel framework that integrates pre-trained generative models with domain-specific fine-tuning to create panoramic images specifically tailored for VLN tasks. This approach overcomes generic generation limitations by enhancing the diversity, relevance, and domain alignment of training environments for VLN applications.
	\item We validated the effectiveness of our framework through two environment generation settings, i.e., masked image inpainting and recursive image outpainting to create entirely new environments with varied room styles, layouts, structures, and objects, thus improving agents' ability to generalize to unseen environments.
	\item Extensive experiments on three widely-used VLN datasets demonstrated that our PanoGen++ outperforms other baselines in terms of navigation error, success rate, and success penalized by path length.
\end{itemize}

The remainder of this paper is organized as follows. 
Section~\ref{sec:Related Work} reviews the related work. 
In Section~\ref{sec:PanoGen++}, we describe the framework of PanoGen++, which generates panoramic environments tailored for VLN and utilizes these environments for agents' pre-training and fine-tuning. 
Section~\ref{sec:Experiments} details the implementation and presents the experimental results. 
Finally, 
conclusions are drawn from our research in Section~\ref{sec:Conclusion}.

\section{Related Work}
\label{sec:Related Work}
Our work is most related to three streams of research, i.e., vision-and-language navigation (VLN), environment augmentation in VLN, and text-to-image generation. 
This section first reviews and summarizes the studies related to these three areas, followed by a comparison of our contributions to existing works.

\subsection{Vision-and-Language Navigation}
Vision-and-language navigation~\cite{anderson2018vision,yue2024safe, wang2024vision, zhou2024navgpt, liu2024volumetric} aims to guide an agent to a target destination based on user instructions in a room environment. Initial work of Anderson \textit{et al.}~\cite{anderson2018vision} introduces a sequence-to-sequence long short-term memory (LSTM) baseline for generating navigational actions based on linguistic and visual inputs. Subsequently, Hao \textit{et al.}~\cite{hao2020prevalent} employed image-text-action triplets to pre-train a multi-modal transformer, facilitating the domain knowledge exploration in VLN. To incorporate historical data, the history-aware multi-modal transformer (HAMT)~\cite{chen2021history} integrates long-term history into multi-modal decision-making processes. Other approaches, such as those proposed by Wang \textit{et al.}~\cite{wang2021structured} and 
Jain \textit{et al.}~\cite{jain2019stay}, construct graphs that embed structured information from instructions and environmental observations, thereby providing explicit semantic relations to guide navigation. Nonetheless, these approaches often simplify visual features in topological maps, which can constrain the detailed grounding of objects and scenes.
The dual-scale graph transformer (DUET)~\cite{chen2022think} addresses these limitations by incorporating both coarse-scale map and fine-scale location embeddings, facilitating efficient global action planning in VLN. More recently, several works have started to exploit the reasoning ability of large language models for navigation tasks. For instance, NavGPT~\cite{zhou2024navgpt} and MapGPT~\cite{chen2024mapgpt} showcase zero-shot navigation in VLN. In particular, NavGPT employs GPT-4 to generate actions, while MapGPT transforms topological maps into exploration cues. To enhance the generalization performance of VLN agents, our approach augments existing VLN models through data augmentation with additional generated environments.

\subsection{Environment Augmentation in VLN}
A primary constraint in VLN is the scarcity of training data, particularly in 3D room environments. This limitation hinders agents' capacity to generalize to unseen scenarios. 
Certain studies focus on eliminating environmental bias or augmenting existing environments.
For instance, CLEAR~\cite{li2022clear} and EnvDrop~\cite{tan2019learning} aim to develop generalized navigation models by learning environment-agnostic representations and utilizing environmental dropout techniques, respectively.
EnvEdit~\cite{li2022envedit} employs generative adversarial networks (GANs)~\cite{goodfellow2020generative} to edit seen environments in terms of style and appearance.
Despite these advancements, these studies remain limited to the existing Matterport3D environments.
Another research aims to introduce new indoor environments from other sources.
Initially, Airbert~\cite{guhur2021airbert} compiles image-caption pairs from online marketplaces, accumulating 1,400 thousand indoor images with 700 thousand captions.
Subsequently, MARVAL~\cite{kamath2023new} constructs navigation trajectories through over 500 new room environments, while AutoVLN~\cite{chen2022learning} generates a large-scale dataset from 900 Habitat-Matterport 3D dataset (HM3D)~\cite{ramakrishnan2habitat} with automatically generated instructions.
Furthermore, ScaleVLN~\cite{wang2023scaling} applies over 1,200 unannotated 3D scans from the HM3D dataset and Gibson~\cite{xia2018gibson} dataset to effectively leverage the augmented data to pre-train and fine-tune an agent. 
However, the collection and processing of such large-scale datasets are resource-intensive and costly, presenting challenges for further scaling these approaches.
To address these issues, PanoGen~\cite{li2023panogen} generates new panoramic environments using Stable Diffusion model
in a recursive image outpainting manner. 
Nevertheless, the panoramic environment generation module of PanoGen was trained on general internet data and is not specifically tailored to VLN environments. Our proposed framework adapts the Stable Diffusion model to create environments more suited for VLN training, thereby enhancing the agents' generalization capabilities.
\subsection{Text-to-Image Generation} 
Text-to-image (T2I) generation~\cite{xu2018attngan, nichol2022glide, rombach2022high,luo2022dualg,zhang2023adding} is a task that takes textual input and produces corresponding images.
Initially, Reed \textit{et al.} introduced the initial GAN-based framework \cite{goodfellow2020generative} for T2I generation.
Building upon this, AttnGAN~\cite{xu2018attngan} enhanced fine-grained detail synthesis by employing attention mechanisms to focus on relevant words within the description. 
DM-GAN~\cite{zhu2019dm} further refined image quality by integrating GANs with a dynamic memory bank. 
Recently, the focus has shifted towards diffusion model-based approaches for T2I generation.
For instance, GLIDE~\cite{nichol2022glide} leverages guided diffusion to produce more photorealistic images using contrastive language-image pre-training (CLIP)~\cite{radford2021learning} guidance and classifier-free guidance. 
Despite their advantages, optimizing these models in pixel space results in low inference speed and high training costs. 
To address these limitations, the latent diffusion model (LDM)~\cite{rombach2022high} operates the diffusion process within a compressed latent space of lower dimension, achieving computational efficiency while maintaining high generation quality.
These methods have been widely employed to support various downstream vision tasks, demonstrating their utility across diverse domains. For instance, T2I techniques have been used in object detection~\cite{lee2024text} and segmentation~\cite{li2022bigdatasetgan} to generate high-quality synthetic training data, augmenting real-world datasets and improving model generalization, particularly in data-scarce scenarios. In our work, the generation module of our proposed PanoGen++ utilizes a pre-trained diffusion model, specifically Stable Diffusion, to create visual environments based on textual input. Additionally, we propose masked image inpainting and recursive image outpainting techniques to generate VLN-specific environments guided by textual descriptions.

\subsection{Positioning of Our Work}
Among the studies discussed above, our PanoGen++ aims to enhance existing VLN training protocols through the application of text-to-image generation techniques for environmental augmentation. 
The most closely works to ours are EnvEdit~\cite{li2022envedit} and PanoGen~\cite{li2023panogen}.  
However, the first method only utilizes the VLN environments for training; while the second method falls short in incorporating the domain-specific knowledge essential for VLN tasks, such as intricate spatial awareness and the distinct characteristics of room scenes.
To meet these specialized requirements, we employ low-rank adaptation to tame the Stable Diffusion to domain-specific environment generation to VLN tasks.
Our approach differs from prior environment augmentation methods in three key aspects:
(i) PanoGen++ employs low-rank matrices to adapt pre-trained diffusion model parameters, significantly reducing the number of learnable parameters and improving computational efficiency;
(ii) by leveraging the extensive world knowledge embedded in large-scale pre-trained diffusion models, while simultaneously focusing on VLN task-specific optimization. PanoGen++ achieves a balance between broad applicability and detailed task performance; 
and (iii) our method integrates inpainting and outpainting techniques to generate new environments, thereby enhancing the diversity and realism of the training scenarios.

\section{PanoGen++}
\label{sec:PanoGen++} 
In this section, we initially outline the problem setup of VLN in Section~\ref{sec:Problem}, followed by the construction room environment descriptions from panoramic view in Section~\ref{sec:text_gen}.
To develop a domain-specific environment generation module for VLN, we subsequently align the generative distribution of PanoGen with that of VLN. 
In Section~\ref{sec:env_gen}, we employ masked image inpainting and recursive image outpainting settings to generate consistent and new panoramic environments. 
These methods are utilized during VLN pre-training and fine-tuning phases to mitigate agent overfitting to training environments. Lastly, the training process for PanoGen++ is elaborated upon.

\subsection{Problem Setup}
\label{sec:Problem}
In VLN, the agent traverses a 3D environment to reach a target location based on natural language instructions.
These instructions consist of a sequence of ${L}$ words, represented as $\mathcal{W} =\left\{w_{0}, \cdots, w_{L-1}\right\}$. The environment is modeled as a environment graph $\mathcal{G}=\{\mathcal{V}, \mathcal{E}\}$, where $\mathcal{V}=\{V_i\}_{i=1}^{K}$ denotes $K$ navigable nodes, and $\mathcal{E}$ represents the connectivity edges.
The agent is equipped with a camera to observe its surroundings and a GPS sensor to record its location.
It starts at an initial node with an initial state $ s_0 $ and receives a panorama $\mathcal{O}_t$ at each time step $t$.
In particular, the panorama $\mathcal{O}_t$ is divided into $N$ images $\mathcal{O}_t=\{o_i\}_{i=1}^{N}$, where $N$ is set to $36$.
At each time step $t$, the agent chooses an action from the action space $\mathcal{A}_{t}$, which includes navigating to a neighboring node 
$V_i \in \mathscr{N}(V_t)$
or stopping at $V_t$.
Once the agent decides to stop at a location, the navigation episode is completed.

\subsection{Aligning Environment Generation Module with VLN}
\label{sec:text_gen}
\textbf{VLN-Specific Environment Training Data Construction.} 
To adapt the pre-trained generation models to VLN-specific environments, we compile a substantial dataset of text and image pairs within VLN contexts.
Initially, we gather room images from the training environments in the room-to-room (R2R)~\cite{anderson2018vision} dataset, which includes 7,705 panoramic images from VLN settings.
Each panoramic observation of an indoor room features numerous objects in a complex arrangement, making it challenging to encapsulate all details in a brief description.
To address this, we divide each 360-degree panorama into 36 discrete sub-view images. 
This partitioning is achieved through a systematic arrangement of 12 horizontal headings and three vertical elevations, with each adjacent view separated by 30-degree intervals.
This results in a total of 277,380 view images, providing a fine-grained representation of the panoramic environment.
We then employ bootstrapping language-image pre-training (BLIP-2) \cite{li2023blip} to caption these segmented sub-view images using the prompt ``a photo of,'' ensuring that each image is matched with a specific description.
Formally, this results in a substantial dataset $\mathcal{D}_{\text{VLN}} = \{ (\mathbf{x}_i, \mathbf{y}_i) \}_{i=1}^{M}$, where $\mathbf{x}_i$ denotes a sub-view image and $\mathbf{y}_i$ represents its corresponding text description.
These domain-specific image-text pairs enhance the performance of the generation model for VLN environment generation by bridging the distribution gap between the pre-trained model and the VLN environment.

\begin{figure*}[!]
    \centering
    \includegraphics[width=0.95\textwidth]{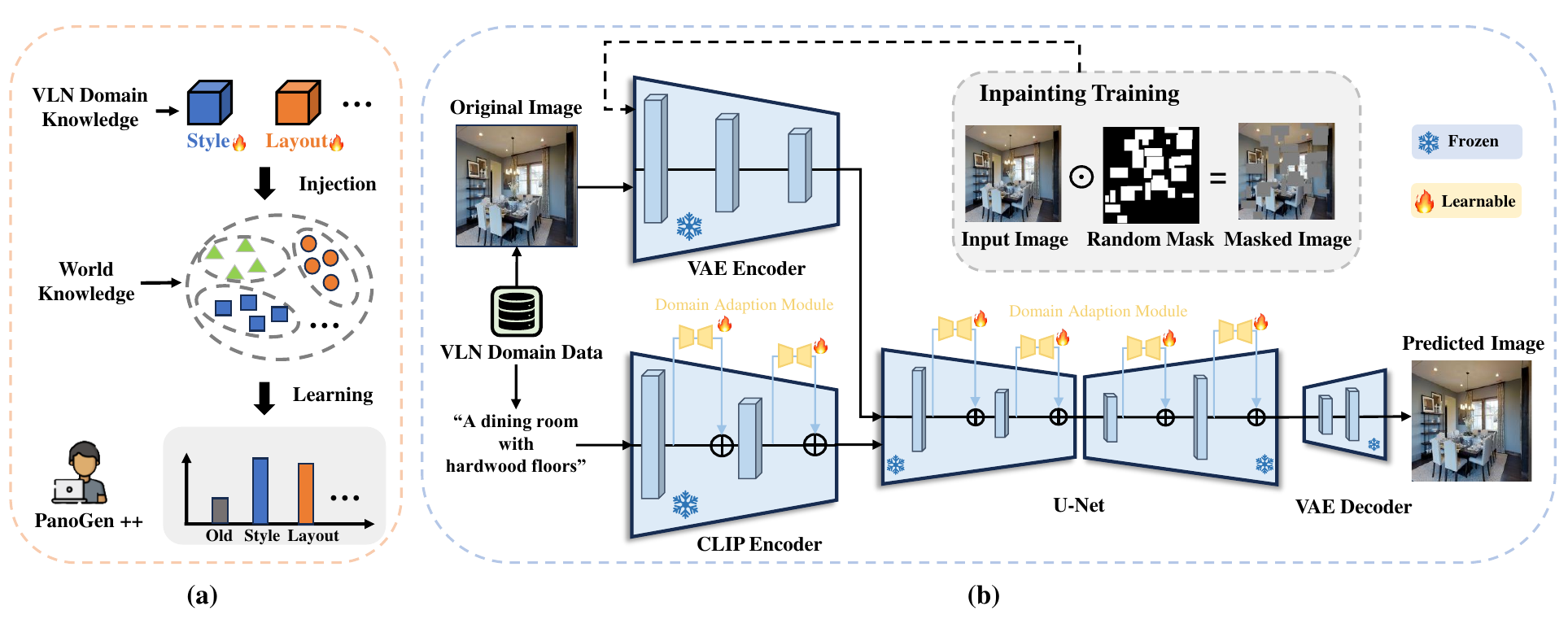}
    \caption{Illustration of domain adaptation for environment generation with VLN. (a) Motivation for adapting PanoGen++ to VLN environments and (b) training pipeline of PanoGen++ (here, given an image from the original VLN environments, it is first encoded into the latent space, following LDM~\cite{rombach2022high}. The environments are then augmented using our generation module, exemplified by the inpainting process, which requires two additional inputs: a mask and a masked image. During training, the VAE and U-Net weights are frozen, and trainable adaption modules are added).}
    \label{fig:method_lora}

\end{figure*}

\textbf{Distribution Alignment for VLN Environment Generation.}
Following the collection of text-image pairs for VLN environments, we utilize the low-rank adaptation (LoRA)~\cite{hu2022lora,kim2024hydra} technique to efficiently adapt the pre-trained generation model to VLN tasks. 
Figure~\ref{fig:method_lora} (a) illustrates the motivation behind PanoGen++, which is to adapt a pre-trained generative model to the VLN domain-specific environment generation. This adaptation enhances the diversity of styles and layouts within the simulation environments of VLN.
This adaptation is achieved by training on the domain-specific dataset, thereby enhancing cross-domain generalization performance.
As depicted in Figure~\ref{fig:method_lora} (b), our domain adaption module optimizes the rank-decomposition matrices of dense layers in diffusion models, making it more parameter-efficient than updating all the parameters of large diffusion models.
In particular, the learnable incremental weight, denoted as $\Delta \mathbf{W}$, to the pre-trained and frozen weights $\mathbf{W}_0 \in \mathbb{R}^{m \times n}\) during fine-tuning exhibits a ``low intrinsic rank.'' 
Consequently, the update $\Delta \mathbf{W}$ can be decomposed into two low-rank matrices, $\mathbf{B} \in \mathbb{R}^{m \times r}$ and $\mathbf{A} \in \mathbb{R}^{r \times n}$, such that $\Delta \mathbf{W} = \mathbf{B}\mathbf{A}$. 
Here, $r$ represents the intrinsic rank of $\Delta \mathbf{W}$, with $r \ll \min(m,n)$.  
During training, only $\mathbf{A}$ and $\mathbf{B}$ are updated to find the suitable $\Delta \mathbf{W} = \mathbf{B}\mathbf{A}$, while keeping $\mathbf{W}_0$ constant. 
For inference, the updated weight matrix $\mathbf{W}$ is obtained as $\mathbf{W} = \mathbf{W}_0 + \mathbf{B}\mathbf{A}$.

Subsequently, we incorporate weights $\Delta \mathbf{W}$ into the text encoder ${\rm E}_{\text{txt}}$ of CLIP~\cite{radford2021learning} model,
and the U-Net from the pre-trained diffusion model.
In our implementation, we use the dataset $\mathcal{D}_{\text{VLN}} = \{ (\mathbf{x}_i, \mathbf{y}_i) \}_{i=1}^{M} $ constructed earlier to train the low-rank matrix $\Delta \mathbf{W}$.
The attention layers in the U-Net are tuned using $\Delta \mathbf{W}$.
These layers focus on adapting only the attention weight matrices in the self-attention module.
Each of these matrices undergoes an affine transformation, defined as:
\begin{equation}
    \label{eq:lora_attn}
    h_{l} = \mathbf{W}_{0, l-1} h_{l-1} + \Delta \mathbf{W}_l h_{l-1},
\end{equation}
where $h$ represents the output of the $l$-th layer of the pre-trained diffusion model.
All other model parameters remain frozen, while $\Delta \mathbf{W}$ is optimized using gradient descent.
Following the training protocol of the LDM~\cite{rombach2022high}, the image $\mathbf{x}$ is encoded by the image encoder ${\rm E}_{\text{img}}$ of variational autoencoder (VAE)~\cite{Kingma2014}
to latent representation $\mathbf{z}$, i.e., $\mathbf{z} = {\rm E}_{\text{img}}(\mathbf{x})$. 
The training objective for domain adaption is described as follows:
\begin{equation}
\label{eq:loss_lora}
    \mathcal{L}_{\text{da}} = \mathbb{E}_{(\mathbf{x}, \mathbf{y}) \sim \mathcal{D}_{\text{VLN}}, \, \epsilon \sim \mathcal{N}(0, {\rm I}), t} \left\| \epsilon - \epsilon_{{\boldsymbol \theta}} (\mathbf{z}_t, \mathrm{E}_{\text{txt}}(\mathbf{y}), t) \right\|_2^2,
\end{equation}
where $\epsilon$ represents noise sampled from a standard Gaussian distribution, i.e., $\epsilon \sim \mathcal{N}(0, {\rm I})\); $t$ is a sampled time step, $\mathbf{z}_t$ is noised latent after the $t$-th time step, 
${\boldsymbol \theta}$ represents the parameters of the LDM, and $\epsilon_{{\boldsymbol \theta}}$ is the estimated errors based on ${\boldsymbol \theta}$.
It is worth noting that only the weights of our domain adaption module are learnable.
After gradient descent optimization of the above loss function, PanoGen++ is capable of generating visual images that conform to the VLN environment distribution while preserving the original world knowledge.

\subsection{Environment Augmentation for VLN Training}
\label{sec:env_gen}
\begin{figure*}[t]
    \centering
    \includegraphics[width=1\textwidth]{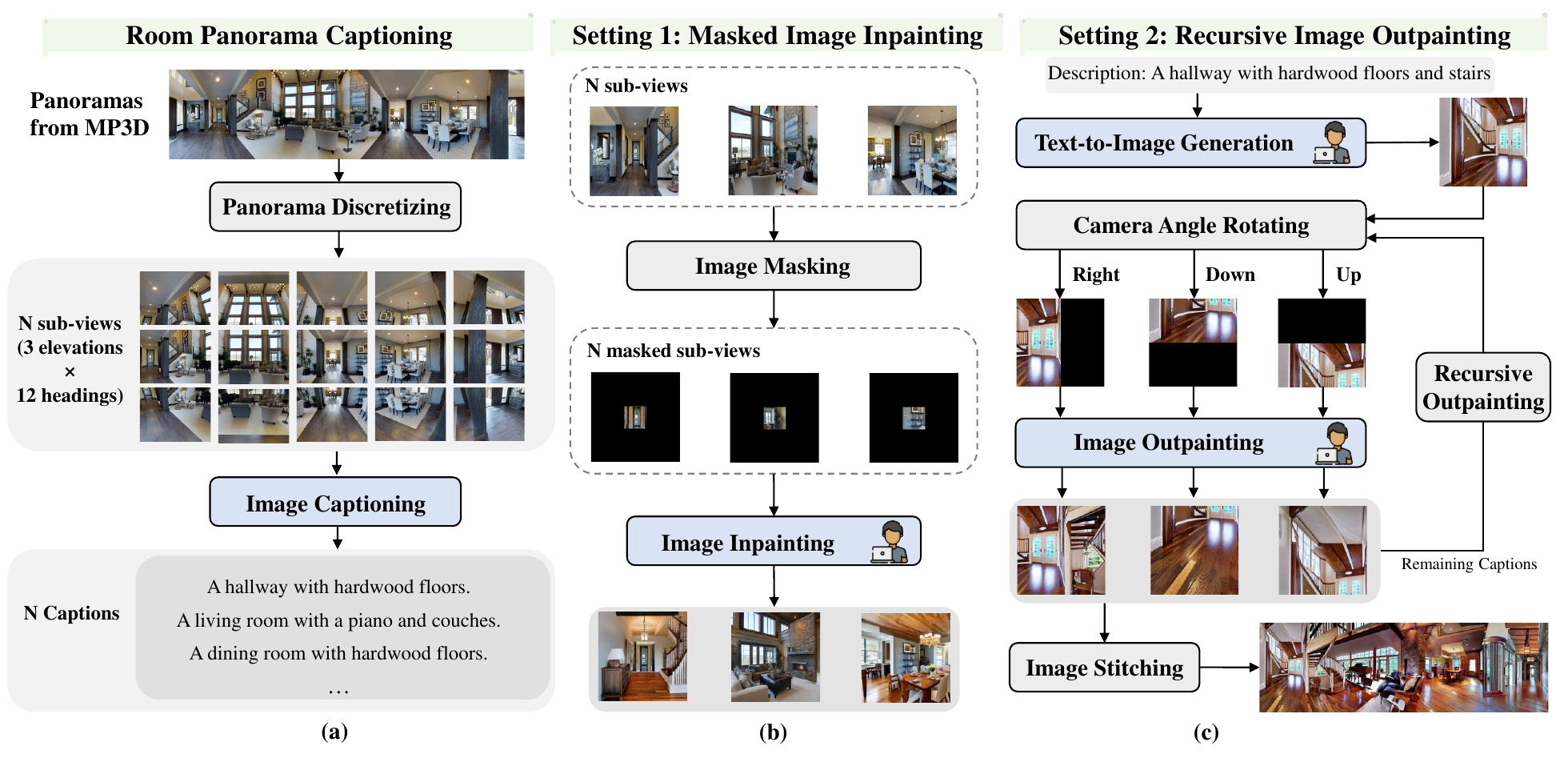}
    \caption{Overview of VLN agent training utilizing PanoGen++ for enhanced environment augmentation.  (a) Room panorama captioning for domain-specific adaptation. (b) Masked image inpainting for semantic alignment. (c) Recursive image outpainting for coherent panoramic extension. (Here, settings for recursive image outpainting follow PanoGen \cite{li2023panogen} for fair comparison, with the addition of a novel masked image inpainting technique for enhanced environment augmentation.)}
    \label{fig:method_panogen++}

\end{figure*}

\textbf{VLN Environment Augmentation.}
Following the distribution alignment of the generation module, PanoGen++ is employed to augment the VLN environment.
As depicted in Figure~\ref{fig:method_panogen++}, we investigate two distinct settings to utilize PanoGen++ for generating panoramic environments.
In the first setting, we implement a center-crop mask combined with a domain-adapted masked image inpainting strategy. This approach involves preserving the central portion of a sub-view from the original panorama and reconstructing the complementary regions using text-guided masked image inpainting. Initially, we extract a center crop from the sub-view of the original panorama, ensuring the core visual content is maintained. The complementary regions of the sub-view are then marked as masked areas. Textual descriptions corresponding to these masked sub-views are then used to guide the image inpainting process with PanoGen++. This technique ensures visual coherence with the retained central crop and semantic alignment with the provided textual guidance. It effectively preserves original visual elements while incorporating novel, contextually relevant environmental features. In the second setting, we implement a two-phase method with recursive image outpainting for text-to-panorama generation. Initially, PanoGen++ is employed to generate a single sub-view as the starting point of the panorama, guided by a corresponding textual caption. Subsequently, an iterative process is applied, shifting the generation window to the right, down, and up, respectively, based on the already generated image. This window retains 50\% of the already generated image and 50\% of the mask to generate a new sub-view. Similar to the inpainting setting, the generated regions of the panoramic view are then outpainted based on the captions of adjacent sub-views using PanoGen++. This recursive image outpainting process, informed by both textual descriptions and contextual sub-views, produces images with enhanced stylistic and layout consistency. Consequently, this approach facilitates the seamless integration of individually generated images into a coherent panoramic composition, resulting in a comprehensive and visually unified representation of the generated environments.
Both settings are utilized as different methods to augment VLN environments during the agent's pre-training and fine-tuning phases.

\textbf{Pre-training.}
During the pre-training phase, the agent undergoes pre-training utilizing three vision-and-language proxy tasks.
In the masked language modeling (MLM) task, 15\% of the tokens in the instruction sequence $\mathcal{W}$ are randomly masked, resulting in $\mathcal{W}_{\backslash i}$.
The model is then tasked with predicting the word distribution within the trajectory $\mathcal{P}_{\mathcal{T}}$:
\begin{equation}
\label{eq:mlm}
    \mathcal{L}_{\text{mlm}} = - \log p(w_i|\mathcal{W}_{\backslash i}, \mathcal{P}_{\mathcal{T}}).
\end{equation}
For the masked region modeling (MRM) task, $P_i \in \mathbb{R}^{C}$ represents the target class probability for a masked observation $\mathcal{O}_i$ (here, $C$ is set to 1,000, which is the same setting as in~\cite{chen2021history}). The model aims to minimize the Kullback-Leibler (KL) divergence between the target and predicted distributions as follows:
\begin{equation}
\label{eq:mrm}
    \mathcal{L}_{\text{mrm}} = - \sum_{j=1}^{C} P_{i,j} \log \hat{P}_{i,j}.
\end{equation}
In the single action prediction (SAP) task,
the objective is to minimize the negative log probability of the predicted action $a_\tau$:
\begin{equation}
    \mathcal{L}_{\mathrm{sap}} = - \sum\nolimits_{\tau=1}^{{\mathcal{T}}}  \text{log}~p(a_\tau|\mathcal{W}, \mathcal P_{<\tau}),
\end{equation}
where $a_t$ represents the predicted action based on a partial demonstration path $\mathcal P_{<t}$ and instruction $\mathcal{W}$.
The full loss for pre-training is then formulated as:
\begin{equation}
\label{eq:l_pt}
    \mathcal{L}_{\text{pt}} = \lambda_1 \mathcal{L}_{\text{mlm}}+\lambda_2 \mathcal{L}_{\text{mrc}} + \lambda_3 \mathcal{L}_{\text{sap}},
\end{equation}
where $\lambda_1$, $\lambda_2$, and $\lambda_3$ are hyper-parameters used to balance the relative contributions of each loss term.

\textbf{Fine-tuning.}
In this phase, the agent undergoes additional training under the supervision of a pseudo interactive demonstrator (PID) denoted as $\pi$.
At each time step, $\pi$ leverages the environment graph $\mathcal{G}$ to determine the subsequent target node.
During each iteration, the current policy is used to sample a trajectory $\mathcal{P}$, and $\pi$ provides pseudo supervision, formalized as follows:
\begin{equation}
\label{eq:pid}
    \mathcal{L}_{\text{pid}} = \sum\nolimits_{t=1}^{T} -\log p(a^{\pi}_t|\mathcal{W},\mathcal{P}_{<t}).
\end{equation}
Here, $a^{\pi}_t$ represents the pseudo target at step $t$. The original expert demonstrations are integrated with the pseudo demonstrations during policy learning, modulated by a balance factor $\lambda$:
\begin{equation}
\label{eq:l_ft}
    \mathcal{L}_\text{ft} = \lambda \mathcal{L}_{\text{sap}} + \mathcal{L}_{\text{pid}}.
\end{equation}

To enhance the agent's generalization capabilities, VLN panoramic environments are augmented during fine-tuning. In particular, a proportion $p\%$ of the observations in the trajectory are randomly replaced with our generated panoramic environments. This replacement strategy avoids overfitting to the constrained training environments. Since these panoramic environments are crafted by PanoGen++, the semantic content of the panoramic observations remains consistent with the original environments.

\subsection{The Training Process of PanoGen++}
In this subsection, we detail the training process for domain adaptation of the pre-trained generation module (refer to \textbf{Algorithm~\ref{algorithm:lora}}) and elucidate pre-training and fine-tuning procedures for the VLN agent (refer to \textbf{Algorithm~\ref{algorithm:vln_training}}).

\textbf{Algorithm~\ref{algorithm:lora}} describes the domain adaptation procedure for the generation module of PanoGen++.
For each iteration (shown in line 1), a batch of image and correspondent description pairs $\{\mathbf{x}, \mathbf{y}\}$ are sampled from $\mathcal{D}_{\text{VLN}}$ (shown in line 2).  
The VAE encoder $\mathrm{E}_{\text{img}}$ then encodes the images $\mathbf{x}$ into latent representations $\mathbf{z}$ (shown in line 3).
The model parameters ${\boldsymbol \theta}$ are derived by combining pre-trained text-to-image generation model parameters ${\boldsymbol \theta}_{G}$ with domain adaptation module parameters ${\boldsymbol \theta}_{\Delta G}$ (shown in line 4).
A time step $t$ and noise vector $\boldsymbol\epsilon$ are randomly sampled (shown in lines 6-7).  
The algorithm then generates a noisy latent representation $\mathbf{z}_t$ using noise schedule parameters $\bar{\alpha}_t$ (shown in line 7, more details can be found in \cite{ho2020denoising}). 
The domain adaptation loss is computed using the full model's noise prediction (shown in line 8), and ${\boldsymbol \theta}_{\Delta G}$ is updated via gradient descent (shown in line 9).
By iterating these steps for $N_{\Delta G}$ iterations, the domain adaptation module parameters ${\boldsymbol \theta}_{\Delta G}$ are optimized, enhancing the pre-trained text-to-image generation model $G$ to generate panoramic environments tailored for VLN tasks.

\textbf{Algorithm~\ref{algorithm:vln_training}} outlines the VLN agent training procedure, comprising pre-training (shown in lines 1-7) and fine-tuning (shown in lines 8-16) phases.  
During pre-training, the agent learns unimodal and multimodal representations through three proxy tasks: MLM, MRM, and SAP (shown in lines 2-4).
The algorithm computes losses for each task, combines them into a total loss $\mathcal{L}_\text{pt}$ (shown in line 5), and updates ${\boldsymbol \theta}_{\text{agent}}$ via gradient descent (shown in line 6).
The fine-tuning phase employs supervised training using pseudo-interactive demonstration.
For each iteration, the algorithm obtains the current VLN environment $e_{\text{orig}}$ from the original MP3D environment $E_{\text{orig}}$ (shown in line 9). It generates an augmented environment $e_{\text{aug}}$ using the pre-trained text-to-image generation model $G$ and optimized domain adaptation module $\Delta G$ (shown in line 10).
The trajectory $\mathcal{P}_{<\tau}$ is updated by combining the current and augmented environments according to a specified percentage $p\%$ (shown in line 11). 
The algorithm computes the PID loss $\mathcal{L}_{\text{pid}}$ using the policy $\pi$ (shown in line 13). The total fine-tuning loss $\mathcal{L}_{\text{ft}}$ is calculated by combining the SAP and PID losses (shown in line 14). Finally, ${\boldsymbol \theta}_{\text{agent}}$ is updated via gradient descent (shown in lines 15).
By iterating these steps, the agent's parameters ${\boldsymbol \theta}_{\text{agent}}$ are optimized to adapt to both original and augmented environments, potentially improving the agent's performance in VLN tasks.
\begin{algorithm}[t]
    \caption{Domain adaptation for the generation module of PanoGen++.}
    \small
    \label{algorithm:lora}
    \SetAlgoLined
    \KwIn{Pre-trained text-to-image generation model $G$, domain adaption module $\Delta G$, image encoder $\mathrm{E}_{\text{img}}$, text encoder $\mathrm{E}_{\text{txt}}$, and dataset $\mathcal{D}_{\text{VLN}} = \left\{ (\mathbf{x}_i, \mathbf{y}_i) \right\}_{i=1}^{N}$.}
    \KwOut{Optimized parameters ${\boldsymbol \theta}_{\Delta G}$ of $\Delta G$.}
    \For{$iter\leftarrow 1$ \KwTo $N_{\Delta G}$}{

        Sample a batch of $\{\mathbf{x}, \mathbf{y}\}$ from $\mathcal{D}_{\text{VLN}}$;
    
        $\mathbf{z} \leftarrow \mathrm{E}_{\text{img}} (\mathbf{x})$\;
        
        ${\boldsymbol \theta} \leftarrow {\boldsymbol \theta}_{G} + {\boldsymbol \theta}_{\Delta G}$\;
        
        $t \sim \operatorname{Uniform}(\{1, \cdots, T\})$\;
        
        $\boldsymbol\epsilon \sim \mathcal{N}(0, \rm I)$\;
        
        $\mathbf{z}_t \leftarrow \sqrt{\bar{\alpha}_{t}} \mathbf{z} + \sqrt{1-\bar{\alpha}_{t}} \boldsymbol{\epsilon}$; \textcolor[rgb]{0.5,0.5,0.5}{\# Forward process of diffusion model, more details can be found in \cite{ho2020denoising}}
        
        $\mathcal{L}_{\text{da}} \gets \mathbb{E}_{(\mathbf{x}, \mathbf{y}) \sim \mathcal{D}_{\text{VLN}}, \, \epsilon \sim \mathcal{N}(0, {\rm I}), t} \left\| \epsilon - \epsilon_{{\boldsymbol \theta}} (\mathbf{z}_t, \mathrm{E}_{\text{txt}}(\mathbf{y}), t) \right\|_2^2$; 
        
        update parameters ${\boldsymbol \theta}_{\Delta G}$ of $\Delta G$ with        \qquad ${\boldsymbol \theta}_{\Delta G} \leftarrow {\boldsymbol \theta}_{\Delta G} - \eta_{\Delta G} \nabla_{{\boldsymbol \theta}_{\Delta G}}\mathcal{L}_\text{da}$\; 
    }
    \Return{${\boldsymbol \theta}_{\Delta G}$;}
\end{algorithm}

\begin{algorithm}[t]
    \caption{VLN agent training with PanoGen++.}
    \small
    \label{algorithm:vln_training}
    \SetAlgoLined
    \KwIn{Instruction-trajectory pair ($\mathcal{W}, \mathcal{P}$), 
    MP3D environment $E_{\text{orig}}$, pre-trained text-to-image generation model $G$, and optimized domain adaption module $\Delta G$.
}
    \KwOut{Optimized parameters ${\boldsymbol \theta}_{\text{agent}}$ of VLN agent.}
  
    \For{$iter\leftarrow 1$ \KwTo $N_{\text{pt}}$}{
        
        $\mathcal{L}_{\text{mlm}} \gets - \log p(w_i|\mathcal{W}_{\backslash i}, \mathcal{P}_{\mathcal{T}})$; \textcolor[rgb]{0.5,0.5,0.5}{\# MLM task}
        
        $\mathcal{L}_{\text{mrm}} \gets - \sum_{j=1}^{C} P_{i,j} \log \hat{P}_{i,j}$; \textcolor[rgb]{0.5,0.5,0.5}{\# MRM task}
        
        $\mathcal{L}_{\mathrm{sap}} \gets - \sum\nolimits_{\tau=1}^{{\mathcal{T}}}  \text{log}~p(a_\tau|\mathcal{W}, \mathcal P_{<\tau})$; \textcolor[rgb]{0.5,0.5,0.5}{\# SAP task}

        $\mathcal{L}_{\text{pt}}\gets \lambda_1 \mathcal{L}_{\text{mlm}}+\lambda_2 \mathcal{L}_{\text{mrm}} + \lambda_3 \mathcal{L}_{\text{sap}}$;  

        update parameters ${\boldsymbol \theta}_{\text{agent}}$  with 
        \qquad  $ {\boldsymbol \theta}_{\text{agent}} \gets {\boldsymbol \theta}_{\text{agent}} - \eta_\text{pt} \nabla_{{\boldsymbol \theta}_{\text{agent}}}\mathcal{L}_{\text{pt}}$; 
    
    }
    
    \For{$iter \gets 1$ \KwTo $N_{\text{ft}}$}{

    $e_{\text{orig}} \gets E_{\text{orig}}(\mathcal{P}_{<\tau})$; \textcolor[rgb]{0.5,0.5,0.5}{\# Obtain current VLN environments}
    
    $e_{\text{aug}} \gets (G+\Delta G)(e_{\text{orig}})$; \textcolor[rgb]{0.5,0.5,0.5}{\# Augment current VLN environments}
    
    $\mathcal{P}_{<\tau} \gets \mathcal{P}_{<\tau-1} + (1-p\%) \cdot e_{\text{orig}} + p\% \cdot e_{\text{aug}}$; \textcolor[rgb]{0.5,0.5,0.5}{\# Mix current and augmented environments}

    $\mathcal{L}_{\mathrm{sap}} \gets - \sum\nolimits_{\tau=1}^{{\mathcal{T}}}  \text{log}~p(a_\tau|\mathcal{W}, \mathcal P_{<\tau})$;

    $\mathcal{L}_{\text{pid}} \gets \sum\nolimits_{t=1}^{T} -\log p(a^{\pi}_t|\mathcal{W},\mathcal{P}_{<t})$; \textcolor[rgb]{0.5,0.5,0.5}{\# Pseudo supervision}

    $\mathcal{L}_{\text{ft}} \gets \lambda \mathcal{L}_{\text{sap}} + \mathcal{L}_{\text{pid}}$;

    update parameters ${\boldsymbol \theta}_{\text{agent}}$ of VLN agent with 
        \qquad ${\boldsymbol \theta}_{\text{agent}} \gets {\boldsymbol \theta}_{\text{agent}}- \eta_\text{ft} \nabla_{{\boldsymbol \theta}_{\text{agent}}}\mathcal{L}_{\text{ft}}$; 
    }
  
\Return{${\boldsymbol \theta}_{\text{agent}}$;}
\end{algorithm}

\section{Experiments}
\label{sec:Experiments}
This section is structured as follows. In Section~\ref{subsec:Implementation Details}, we provide a detailed description of the implementation aspects of the proposed PanoGen++. Following this, in Section~\ref{subsec:Comparisons}, we present both qualitative and quantitative evaluations to empirically validate the superiority of our method compared to existing approaches. To further substantiate our claims, Section~\ref{subsec:Ablation Study} offers an in-depth ablation study, assessing the contributions of the key components within PanoGen++. Finally, Section~\ref{sec:lim_and_dis} discusses the limitations and offers further insights.

\subsection{Implementation Details}
\label{subsec:Implementation Details}

\textbf{Datasets.}
We evaluated our proposed method using three VLN datasets based on the Matterport3D indoor environments: room-to-room (R2R)~\cite{anderson2018vision}, room-for-room (R4R)~\cite{jain2019stay}, and cooperative vision-and-dialog navigation (CVDN)~\cite{thomason2020vision}. 
The training sets for these datasets encompass 61 unique room environments. The unseen validation and test sets, which are not used during training, consist of 11 and 18 room environments, respectively.
In particular, the R2R dataset includes 21,567 trajectory-instruction pairs, with each instruction averaging 32 words and the corresponding ground-truth paths encompassing an average of seven nodes, spanning a mean total distance of 10 meters. R4R increases the complexity by concatenating pairs of paths and their associated instructions, resulting in approximately twice the number of steps and path lengths compared to R2R. The CVDN dataset features ambiguous dialogs between a navigator, who seeks the target through guidance, and an oracle with privileged knowledge of the optimal next step, allowing the agent to conduct and interpret human dialogs to navigate effectively. 

\textbf{Network Training.}
For the domain adaptation of the PanoGen++ generation module, all descriptions are generated within the training environments of the R2R dataset using the BLIP-2 model, specifically the BLIP-2-FlanT5-xxL version.
During the domain adaptation phase, a rank of 64 was employed for the corresponding module.
This stage utilized a constant learning rate $\eta_{\Delta G}$ of $1\times 10^{-7}$, with a batch size of 8 over 40,000 training iterations. 
The generation module was built upon the pre-trained Stable Diffusion v2.1 model and its corresponding inpainting model. To preserve essential visual information in the masked image inpainting configuration, a center crop of 128 $\times$ 128 pixels was implemented.
For the VLN agent training with PanoGen++, we adopted the architecture from DUET~\cite{chen2022think}, adhering to its established hyperparameters. The model was trained on a single A100 GPU. 
The pre-training phase of the VLN agent employed a batch size of 64 for 100,000 iterations. Optimization was conducted using the Adam~\cite{KingBa15} optimizer with $\beta_1 = 0.9$ and $\beta_2 = 0.98$, and a learning rate $\eta_\text{pt}$ of $5\times 10^{-5}$.
The hyper-parameters $\lambda_1$, $\lambda_2$, and $\lambda_3$ in Eq.~(\ref{eq:l_pt}) and $\lambda$ in Eq.~(\ref{eq:l_ft}) were configured according to the implementation of DUET.
The fine-tuning phase utilized a batch size of 8, a learning rate $\eta_\text{ft}$ of $1\times 10^{-5}$, and consisted of 40,000 iterations.

\subsection{Comparisons}
\label{subsec:Comparisons}
\textbf{Evaluation Metrics.}
For both the R2R and R4R datasets, we evaluate performance using four key metrics: (i) trajectory length (TL), which calculates the mean length of the predicted path in meters; (ii) navigation error (NE), denoting the average distance in meters from the agent's final location to the target; (iii) success rate (SR), defined as the proportion of agents that stop within three meters of the target viewpoint; and (iv) success rate weighted by path length (SPL), which penalizes longer paths that deviate from the provided instructions to reach the target. It should be noted here that SR and SPL are the primary metrics for evaluation on the R2R and R4R datasets. For CVDN, goal progress (GP) serves as the sole metric, measuring the average difference between the length of the completed trajectory and the remaining distance to the goal.

\begin{table*}[t]
	\centering
	\renewcommand{\arraystretch}{1.3}
	\tabcolsep=0.2cm

 \caption{Comparison with other baselines on the R2R dataset in terms of trajectory length (TL), navigation error (NE), success rate (SR), and success rate weighted by path length (SPL) (here, for NE, lower values are better, while for SR and SPL, higher values are better).}
 \label{table:table1}
	
	\scalebox{1}{
		\begin{tabular}{lcccccccc} 
			\hline 
			\multicolumn{1}{l}{\multirow{2}{*}{\centering Method}} & \multicolumn{4}{c}{Validation Unseen} & \multicolumn{4}{c}{Test Unseen} \\
			\cline{2-9}
			& TL & NE ($\downarrow$) & SR ($\uparrow$) & SPL ($\uparrow$) & TL & NE ($\downarrow$) & SR ($\uparrow$) & SPL ($\uparrow$) \\
			\hline
			EnvDrop~\cite{tan2019learning} & 10.70 & 5.22 & 52.00 & 48.00 & 11.66 & 5.23 & 51.00 & 47.00  \\
            HAMT~\cite{tan2019learning} & 11.46 & \textbf{2.29} & 66.00 & 61.00 & 12.27 & 3.93 & 65.00 & 60.00 \\
			PREVALENT~\cite{hao2020prevalent} & 10.19 & 4.71 & 58.00 & 53.00 & 10.51 & 5.30 & 54.00 & 51.00 \\
			AirBERT~\cite{guhur2021airbert} & 11.78 & 4.01 & 62.00 & 56.00 & 12.41 & 4.13 & 62.00 & 57.00 \\
			RecBERT~\cite{Hong_2021_CVPR} & 12.01 & 3.93 & 63.00 & 57.00 & 12.35 & 4.09 & 63.00 & 57.00  \\ 
			REM~\cite{liu2021vision} & 12.44 & 3.89 & 63.60 & 57.90 & 13.11 & 3.87 & 65.20 & 59.10  \\ 
			EnvEdit~\cite{li2022envedit}  & 12.13 & 3.22 & 67.90 & 62.90 & -- & -- & -- & --  \\
            DUET~\cite{chen2022think} & 13.94 & 3.31 & 72.00 & 60.00 &  14.73 & 3.65 & 69.00 & 59.00  \\
			MARVAL~\cite{kamath2023new} & 10.15 & 4.06 & 64.80 & 60.70 & 10.22 & 4.18 & 62.00 & 58.00  \\
			SE3DS~\cite{koh2023simple} & -- & 3.29 & 69.00 & 62.00 & -- & 3.67 & 66.00 & 60.00 \\

            AZHP~\cite{gao2023adaptive} & 14.05 & 3.3.15 & 72.00 & 61.00 & 14.95 & 3.52 & 71.00 & 60.00 \\
			PanoGen~\cite{li2023panogen}  & 17.47 & 2.95 & 74.03 & 59.54 & 14.38 & 3.31 & 71.70 & 61.90 \\
			\hline
           PanoGen++ (inpainting setting) & 14.99 & 2.91 & \underline{75.02} & \textbf{65.29} & 15.16 & \textbf{2.86} & \underline{73.26} & \textbf{64.48} \\
           PanoGen++ (outpainting setting) &  15.36 & \underline{2.88} & \textbf{75.12} & \underline{64.88} & 15.40 & \underline{2.94} & \textbf{74.14} & \underline{63.98} \\
			\hline
	\end{tabular}}
\end{table*}

\textbf{Baselines.}
To assess the efficacy of PanoGen++, we conducted a comparative analysis against several baselines. 
These baselines can be classified into two primary categories: environment augmentation-based and representation learning-based methods.
The first category encompasses EnvDrop~\cite{tan2019learning}, AirBERT~\cite{guhur2021airbert}, REM~\cite{liu2021vision}, EnvEdit~\cite{li2022envedit}, SE3DS~\cite{koh2023simple}, and PanoGen~\cite{li2023panogen}. These methods focus on enhancing the training environments to improve navigation performance.
The second category comprises state-of-the-art representation learning-based methods, including PREVALENT~\cite{hao2020prevalent}, RecBERT~\cite{Hong_2021_CVPR}, HAMT~\cite{tan2019learning}, MARVAL~\cite{kamath2023new}, and DUET~\cite{chen2022think}. These approaches emphasize the importance of model pre-training in VLN.
To maintain experimental integrity, we implemented all baseline methods using the original source code when available. In cases where implementation details or code were not fully accessible, we utilized the results reported in the respective publications.
Given that the primary objective of the VLN task is to navigate successfully in unseen environments, we report performance metrics based on the highest success rate achieved on the validation unseen dataset.

\begin{figure}[t]
  \centering
  \includegraphics[width=1.0\columnwidth]{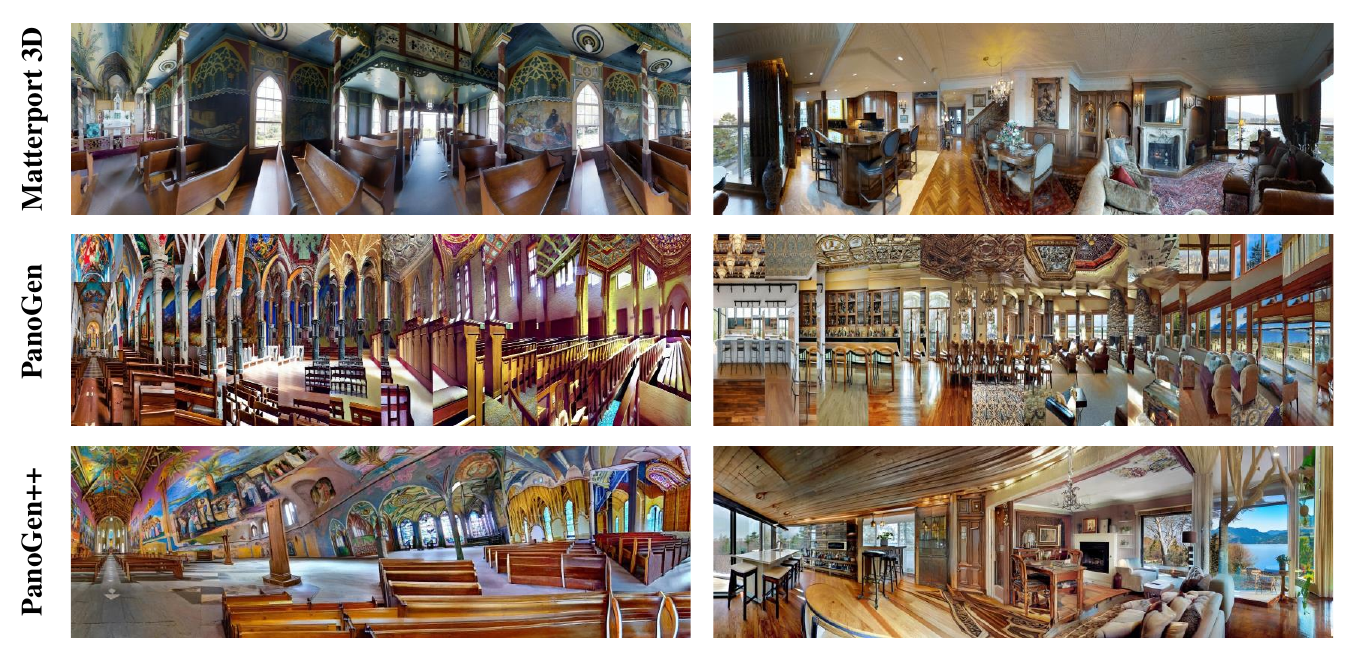}
  \caption{Qualitative analysis of the panoramic environments generated by our PanoGen++. Here, the Matterport3D serves as the original environment for VLN tasks, while PanoGen represents the panoramic environments generated by the PanoGen.}
  \label{fig:result}

\end{figure}

\begin{figure}[t]
  \centering
  \includegraphics[width=1\columnwidth]{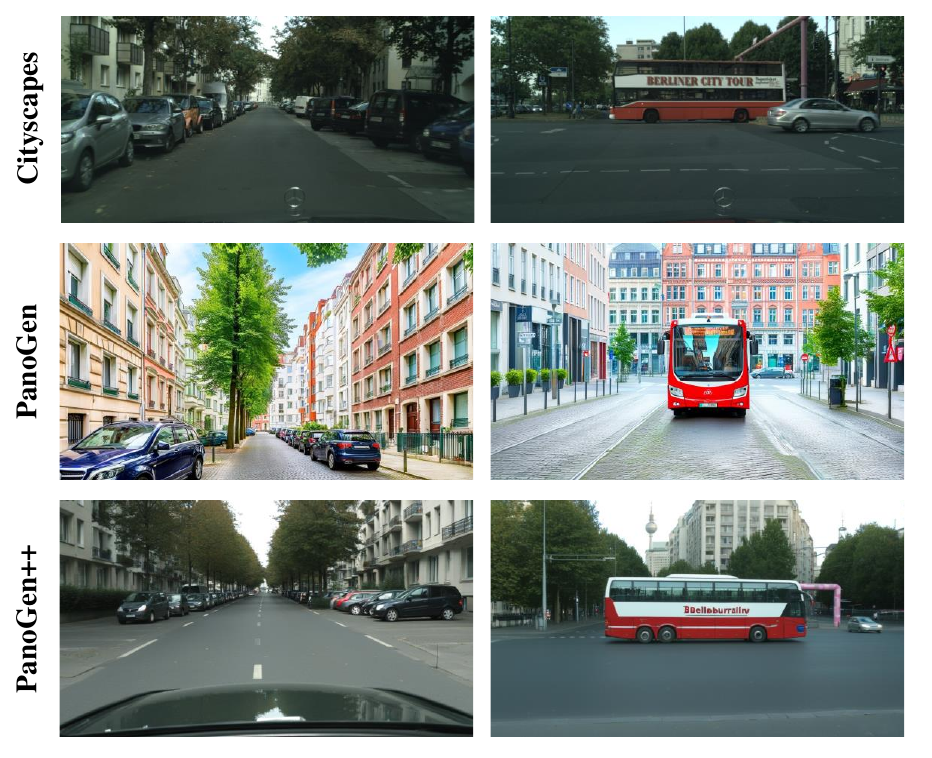}
  \caption{Qualitative analysis of the panoramic environments generated on the Cityscapes dataset. Here, the original Cityscapes images serve as the reference, while PanoGen++ and PanoGen represent the generated panoramic environments by our method and the baseline respectively.}
  \label{fig:city}

\end{figure}


\begin{table}[t]
    \centering
    \caption{Comparison of PanoGen++ and PanoGen on visual authenticity (FID) across Matterport3D (a VLN simulation environment) and Cityscapes (a real-world urban scenario). Lower FID scores indicate smaller domain gaps.}
    \label{table:fid}
    \scalebox{0.9}{
    \begin{tabular}{l c c}
        \toprule
        \multirow{2}{*}{Method} & \multicolumn{2}{c}{FID ($\downarrow$)} \\
         & Matterport3D & Cityscapes \\
        \midrule
        PanoGen~\cite{li2023panogen} & 40.754 & 38.238 \\
        PanoGen++ (inpainting setting) & 32.908 & 22.156 \\
        PanoGen++ (outpainting setting) & 38.081 & 27.843 \\
        \bottomrule
    \end{tabular}}
\end{table}


\textbf{Qualitative Comparison.} 
As illustrated in Figure~\ref{fig:result}, we present several panoramic environments generated using PanoGen++ in comparison to those produced by PanoGen. 
The proposed method demonstrates superior capability in capturing intricate details, specifically tailored for VLN tasks.
For instance, in the original panorama, PanoGen++ successfully replicates the church's mural details, including the palm tree on the left side, whereas PanoGen fails to reproduce these nuanced elements.
The environments generated by PanoGen++ exhibit high visual fidelity, closely approximating real-world settings. This realism is essential for the effective training of navigation agents, providing diverse and authentic scenes for data augmentation. Furthermore, the recursive outpainting approach employed by PanoGen++ generates continuous views that can be seamlessly stitched into high-quality panoramas while preserving the original layout, as evidenced in the rightmost panorama.
To further validate the effectiveness of PanoGen++, the Fréchet inception distance (FID) was utilized.
FID quantifies the similarity between feature statistics of synthesized and real images, with lower values indicating closer distribution alignment between two domains.
The FID metric was calculated on a subset of 10,000 images from both PanoGen and PanoGen++. 
Table~\ref{table:fid} demonstrates that PanoGen++ significantly outperforms PanoGen in visual authenticity.
Specifically, in the Matterport3D setting, PanoGen++ demonstrates significant improvements in FID scores, with the inpainting and outpainting configurations achieving enhancements of approximately 7.846 and 2.673, respectively, compared to PanoGen.
The observed performance difference can be attributed to three key factors. First, the inpainting setting aligns more closely with the model's core training objective, as it retains direct access to ground-truth visual features through the preserved center crop. Second, the recursive generation process in the outpainting setting, although an improvement over PanoGen, presents additional challenges in maintaining global consistency. Third, the effective incorporation of VLN domain-specific knowledge into the pre-trained diffusion model significantly enhances the realism and quality of the generated images across both settings.
To validate the suitability of PanoGen++ for the VLN corpus without considering other benchmarks or real-world scenarios, we conducted additional evaluations using the Cityscapes dataset \cite{cordts2016cityscapes}. As illustrated in Figure \ref{fig:city}, PanoGen++ generates images that not only align with the stylistic attributes of the Cityscapes domain but also introduce coherent and realistic variations. In contrast, while PanoGen’s outputs are structurally sound, they lack the fidelity and domain-specific nuances captured by our parameter-efficient fine-tuning approach. Furthermore, as shown in Table~\ref{table:fid}, even without leveraging other real-world datasets (e.g., Cityscapes) during training, PanoGen++ consistently demonstrates strong visual authenticity across diverse datasets.

\textbf{Quantitative Comparison.}  
Tables~\ref{table:table1} and~\ref{table:table2} present comprehensive performance comparisons of state-of-the-art methods on the R2R, R4R, and CVDN datasets, showcasing the advancements achieved by our proposed method, PanoGen++. 
The results suggest that PanoGen++ consistently exhibits superior performance across multiple metrics and datasets, surpassing both the latest data enhancement method, PanoGen, and the leading representation learning-based method, DUET.
In particular, on the R2R benchmark (Table~\ref{table:table1}), PanoGen++ outperforms all other baselines in primary metrics (SR and SPL) across both unseen validation and test splits, highlighting its robustness and generalizability. In the masked image inpainting setting, PanoGen++ attains the best performance on the test unseen split with an SR of 73.47\% and an SPL of 62.05\%, improving upon PanoGen by 1.77\% and 0.15\% respectively. In comparisonto DUET, PanoGen++ demonstrates even more substantial gains, with a 3.27\% improvement in SR and a 2.55\% increase in SPL. These results emphasize the superior generalization capabilities of our method in truly unseen environments.
In the recursive image outpainting setting, PanoGen++ achieves the highest SR of 74.67\% on the validation unseen split, surpassing the previous state-of-the-art environment augmentation method PanoGen by 0.64\%. Notably, PanoGen++ also outperforms DUET by 2.17\% in SR on the same split, underscoring the effectiveness of our data enhancement approach in comparison to representation learning techniques.
Table~\ref{table:table2} provides further evidence of PanoGen++'s robustness across different VLN tasks. On the R4R dataset, which features longer trajectories and instructions, PanoGen++ significantly outperforms PanoGen, achieving a 2.28\% improvement in SR (from 47.78\% to 50.06\%) and a substantial reduction in NE from 6.02 to 2.37. This demonstrates our method's effectiveness in handling more complex navigation scenarios.
For the CVDN dataset, which evaluates navigation based on dialogue history, PanoGen++ achieves a GP of 6.13, surpassing PanoGen by 0.5 meter (an 8.9\% relative improvement). This substantial gain indicates that our agent has acquired useful commonsense knowledge from the diverse visual environments in PanoGen++, enabling better generalization to unseen environments when navigating based on under-specified instructions in the CVDN dataset.

\begin{table*}[!t]
    \caption{Results of ablation study on the impact of replacing the original environment with our panoramic observation during fine-tuning on the validation unseen set for R2R, R4R, and CVDN datasets.
    }
    \label{table:table2}
    \centering
        \begin{tabular}{l|cccc|cccc|c}
            \toprule
            \multirow{2}{*}{Method} & \multicolumn{4}{c|}{R2R} & \multicolumn{4}{c|}{R4R} & CVDN \\ \cmidrule{2-10}
            & TL & NE ($\downarrow$) & SR ($\uparrow$) & SPL ($\uparrow$) & TL & NE ($\downarrow$) & SR ($\uparrow$) & SPL ($\uparrow$) & GP ($\uparrow$) \\
            \midrule
            DUET~\cite{chen2022think} & 13.94 & 3.31 & 72.00 & 60.00 & -- & -- & -- & -- & -- \\
            PanoGen~\cite{li2023panogen} & 17.47 & 2.95 & 74.03 & 59.54 & 18.62 & 6.02 & 47.78 & 44.25 & 5.63 \\
            PanoGen++ (ours) & 15.36 & \textbf{2.88} & \textbf{75.12} & \textbf{64.88} & 18.35 & \textbf{5.84} & \textbf{48.41} & \textbf{44.93} & \textbf{6.38} \\
        
            \bottomrule
        \end{tabular}

\end{table*}

\subsection{Ablation Study}
\label{subsec:Ablation Study}
In this subsection, three sets of experiments were carried out to investigate the effectiveness of environments augmented by PanoGen++, evaluate the correlation between panoramic environment quantity and agent performance, and examine the domain adaptation module's rank. 
These studies aim to demonstrate how these factors influence the generalizability of agents to unseen environments.
\begin{table}[t]
	\centering
    \caption{Comparison of replacing varying ratios (i.e., $p$\%) of viewpoints in the trajectory with panoramic environments generated by PanoGen++ on the R2R validation unseen set.
    }
    \label{table:table3}
	\begin{tabular}{c *{5}{c}}
		\toprule
		\multirow{2}{*}{No.} & \multirow{2}{*}{Ratio} & \multicolumn{4}{c}{Validation Unseen} \\
		\cmidrule(lr){3-6}
		& & TL & NE ($\downarrow$) & SR ($\uparrow$) & SPL ($\uparrow$) \\
		\midrule
		1 & 0.0 & 13.94 & 3.31 & 72.00 & 60.00 \\
		2 & 0.1 & 15.29 & 2.97 & 73.82 & 62.43 \\
		3 & 0.3 & 15.40 & 2.94 & 74.14 & 63.98 \\
		4 & 0.5 & 15.36 & \textbf{2.88} & \textbf{75.12} & 64.88 \\
		5 & 0.7 & 15.72 & 2.88 & 74.73 & \textbf{65.02} \\
		\bottomrule
	\end{tabular}

\end{table}

\textbf{Effectiveness of Panorama Environment Augmented by PanoGen++.}
To determine whether augmenting environments with PanoGen++ during fine-tuning enhances agents' generalization to unseen environments, we conducted experiments where random observations in the trajectory were replaced with our domain-specific panoramic environments generated by PanoGen++.
As indicated in Table~\ref{table:table2}, this replacement led to improved navigation performance across all three datasets. Specifically, our method resulted in a 0.67\% increase in SR on the R2R dataset, an improvement of 0.5 meter in GP on the CVDN dataset, and enhancements of 2.28\% in SR and 3.65\% in NE on the R4R dataset.
These consistent improvements across all datasets validate the efficacy of PanoGen++ environments.
Moreover, balancing the ratio of replaced observations in the full navigation trajectory proved essential. As shown in Table~\ref{table:table3}, the optimal performance was achieved when 50\% of the viewpoints were replaced, with the highest SPL performance observed at a 70\% replacement rate.
These improvements highlight the high visual quality of PanoGen++ environments, which provide steady performance gains when a significant portion of viewpoints is replaced, demonstrating their suitability for VLN training.

\begin{table}[t]
	\centering
    \caption{Comparison of different masking strategies in inpainting-based environment generation on the R2R validation unseen set.
    }
    \label{table:table4}
    \scalebox{0.9}{
	\begin{tabular}{c *{5}{c}}
		\toprule
		\multirow{2}{*}{No.} & \multirow{2}{*}{Masking Strategy} & \multicolumn{4}{c}{Validation Unseen} \\
		\cmidrule(lr){3-6}
		& & TL & NE ($\downarrow$) & SR ($\uparrow$) & SPL ($\uparrow$) \\
		\midrule
		1 & SRM & 13.63 & 2.96 & 74.50 & 63.91 \\
		2 & ERM & 14.25 & 2.94 & 73.86 & 62.44 \\
		3 & HIM & 13.84 & 3.06 & 73.61 & 63.06 \\
		4 & PRM & 14.99 & \textbf{2.91} & \textbf{75.02} & \textbf{65.29} \\
		\bottomrule
	\end{tabular}}

\end{table}

\textbf{Impact of Different Masking Strategies on Inpainting-Based Environment Generation.}
We systematically investigated four distinct masking strategies within our inpainting-based environment generation framework: (i) sparse region masking (SRM), which employs rectangular masks randomly distributed over 10-50\% of the image area; (ii) extensive region masking (ERM), which extends the rectangular masked regions to cover 50-90\% of the image; (iii) half-image masking (HIM), which implements a deterministic approach by masking half of the image area; and (iv) peripheral region masking (PRM), which masks the peripheral areas while preserving an unmasked central region of the image.
As shown in Table~\ref{table:table4}, PRM demonstrates superior performance across multiple evaluation metrics, achieving a navigation error reduction of 0.22m and 0.14m compared to ERM and HIM, respectively. Regarding success rate, PRM exhibits notable improvements of 0.98\% over ERM and 1.28\% over HIM. While SRM achieves the highest SPL of 63.97\%, the distributed rectangular masking pattern in this approach introduces potential spatial discontinuities that may compromise the structural integrity of the generated environments.
The superior performance of PRM can be attributed to its fundamental characteristics in preserving the most salient visual information that agents primarily utilize for navigation decision-making. By maintaining an unmasked central region, this strategy facilitates smooth and consistent environmental modifications that correspond to the natural progression of visual information during navigation. This approach ensures that generated content seamlessly integrates with existing environmental features while preserving critical architectural and spatial relationships within the scene. The comparative analysis of alternative strategies further substantiates our findings: the extensive masking in ERM potentially disrupts critical spatial relationships, while the rigid partitioning in HIM may introduce artificial boundaries that impede natural visual flow. Based on this comprehensive empirical evaluation, we selected PRM as our primary masking strategy due to its proven ability to achieve an optimal balance between preserving local features and maintaining global spatial coherence. This balance enhances the effectiveness of environment generation for VLN tasks.

\textbf{Impact of Quantity of Panoramic Environments on VLN Training.}
To evaluate the effect of the number of panoramic environments, we randomly selected 2,500, 4,500, and 6,500 panoramic views from the 7,705 available in the R2R training environments to conduct experiments. 
During training of our domain adaption module of PanoGen++, our method effectively captured the intricate nuances and diverse characteristics of the styles presented by these environments.
As illustrated in Table~\ref{table:table5}, fewer panoramic views diminish agent performance due to insufficient domain-specific knowledge acquisition.
Conversely, training with more panoramic environments consistently enhanced performance.
Additionally, generating one more panoramic view for each panorama in the original R2R training environments doubled the number of panoramic environments.
As shown in Table~\ref{table:table5}, the SPL gain continued to increase with additional VLN original environments for training the domain adaption module (No. 6 vs. No. 5), suggesting that incorporating more panoramic environments from the VLN domain can further enhance agents' generalizability to new environments.
\begin{table*}[t]
	\centering
	\renewcommand{\arraystretch}{1.3}
	\tabcolsep=0.2cm
	\caption{Comparison of PanoGen++ and other baselines on the REVERIE dataset, evaluated using trajectory length (TL), oracle success rate (OSR), success rate (SR), success rate weighted by path length (SPL), remote grounding success (RGS), and remote grounding success penalized by path length (RGSPL). Note that, except for TL, higher values indicate better performance across all metrics.
	}
	\label{table:table_reverie}
	\scalebox{0.75}{
		\begin{tabular}{lcccccccccccc} 
			\hline 
			\multicolumn{1}{l}{\multirow{2}{*}{\centering Method}} & \multicolumn{6}{c}{Validation Seen} & \multicolumn{6}{c}{Validation Unseen} \\
			\cline{2-13}
			& TL & OSR ($\uparrow$) & SR ($\uparrow$) & SPL ($\uparrow$) & RGS & RGSPL ($\uparrow$) & TL & OSR ($\uparrow$) & SR ($\uparrow$) & SPL ($\uparrow$) & RGS ($\uparrow$) & RGSPL ($\uparrow$) \\
			\hline
			Seq2Seq~\cite{anderson2018vision} & 12.88 & 35.70 & 29.59 & 24.01 & 18.97 & 14.96 & 11.07 & 8.07 & 4.20 & 2.84 & 2.16 & 1.63 \\
			RCM~\cite{wang2019reinforced} & 10.70 & 29.44 & 23.33 & 21.82 & 16.23 & 15.36 & 11.98 & 14.23 & 9.29 & 6.97 & 4.89 & 3.89 \\
			SMNA~\cite{ma2019selfmonitoring} & 7.54 & 43.29 & 41.25 & 39.61 & 30.07 & 28.98 & 9.07 & 11.28 & 8.15 & 6.44 & 4.54 & 3.61 \\
			FAST-MATTN~\cite{qi2020reverie} & 16.35 & 55.17 & 50.53 & 45.50 & 31.97 & 29.66 & 45.28 & 28.20 & 14.40 & 7.19 & 7.84 & 4.67 \\
			SIA~\cite{lin2021scene} & 13.61 & 65.85 & 61.91 & 57.08 & 45.96 & 42.65 & 41.53 & 44.67 & 31.53 & 16.28 & 22.41 & 11.56 \\
			RecBERT~\cite{Hong_2021_CVPR} & 13.44 & 53.90 & 51.79 & 47.96 & 38.23 & 35.61 & 16.78 & 35.02 & 30.67 & 24.90 & 18.77 & 15.27 \\
			Airbert~\cite{guhur2021airbert} & 15.16 & 48.98 & 47.01 & 42.34 & 32.75 & 30.01 & 18.71 & 34.51 & 27.89 & 21.88 & 18.23 & 14.18 \\
			HAMT~\cite{chen2021history} & 12.79 & 47.65 & 43.29 & 40.19 & 27.20 & 25.18 & 14.08 & 36.84 & 32.95 & 30.20 & 18.92 & 17.28 \\
			DUET~\cite{chen2022think} & 13.86 & \underline{73.86} & \underline{71.75} & \underline{63.94} & \underline{57.41} & \underline{51.14} & 22.11 & \underline{51.07} & \underline{46.98} & \underline{33.73} & \underline{32.15} & \underline{23.03} \\
			AZHP~\cite{gao2023adaptive} & 13.95 & \textbf{75.12} & \textbf{74.14} & \textbf{67.22} & \textbf{59.80} & \textbf{54.20} & 22.32 & \textbf{53.65} & \textbf{48.31} & \textbf{36.63} & \textbf{34.00} & \textbf{25.79} \\
			CKR~\cite{gao2024room} & 12.16 & 61.91 & 57.27 & 53.57 & -- & -- & 26.26 & 31.44 & 19.14 & 11.84 & -- & --  \\
			PanoGen~\cite{li2023panogen} & 14.97 & 58.12 & 60.15 & 49.08 &  44.48 & 37.72 & 20.35 & 46.46 & 43.34 & 31.74 & 28.34 & 20.88 \\
			PanoGen++ (inpainting setting) & 14.50 & 63.46 & 66.97 & 54.73 & 50.11 & 43.54 & 21.09 & 50.30 & 45.47 & 32.03 & 29.68 & 21.17 \\
			PanoGen++ (outpainting setting) & 16.49 & 62.75 & 66.41 & 51.70 & 48.77 & 40.46 & 23.15 & 50.87 & 45.13 & 30.39 & 29.31 & 19.83 \\
			
			\hline
	\end{tabular}}
\end{table*}

\begin{table}[htbpt]
	\centering
    \caption{Comparison of different numbers of panoramic views used in training the domain adaptation module of PanoGen++.
    }
    \label{table:table5}
	\begin{tabular}{c *{5}{c}}
		\toprule
		\multirow{2}{*}{No.} & \multirow{2}{*}{\# Views} & \multicolumn{4}{c}{Validation Unseen} \\
		\cmidrule(lr){3-6}
		& & TL & NE ($\downarrow$) & SR ($\uparrow$) & SPL ($\uparrow$) \\
		\midrule
		1 & 0 & 17.47 & 2.95 & 74.03 & 59.54 \\
		2 & 1,000 & 14.53 & 3.08 & 73.48 & 62.47 \\
		3 & 2,500 & 15.54 & 2.97 & 74.24 & 63.74 \\
		4 & 4,500 & 15.49 & 2.92 & 74.63 & 64.00 \\
		5 & 6,500 & 15.36 & \textbf{2.88} & \textbf{75.12} & \textbf{64.88} \\
		6 & 7,705 & 15.49 & 2.90 & 74.73 & 64.29 \\
		\bottomrule
	\end{tabular}
\end{table}

\begin{figure}[t]
  \centering
\includegraphics[width=0.5\textwidth]{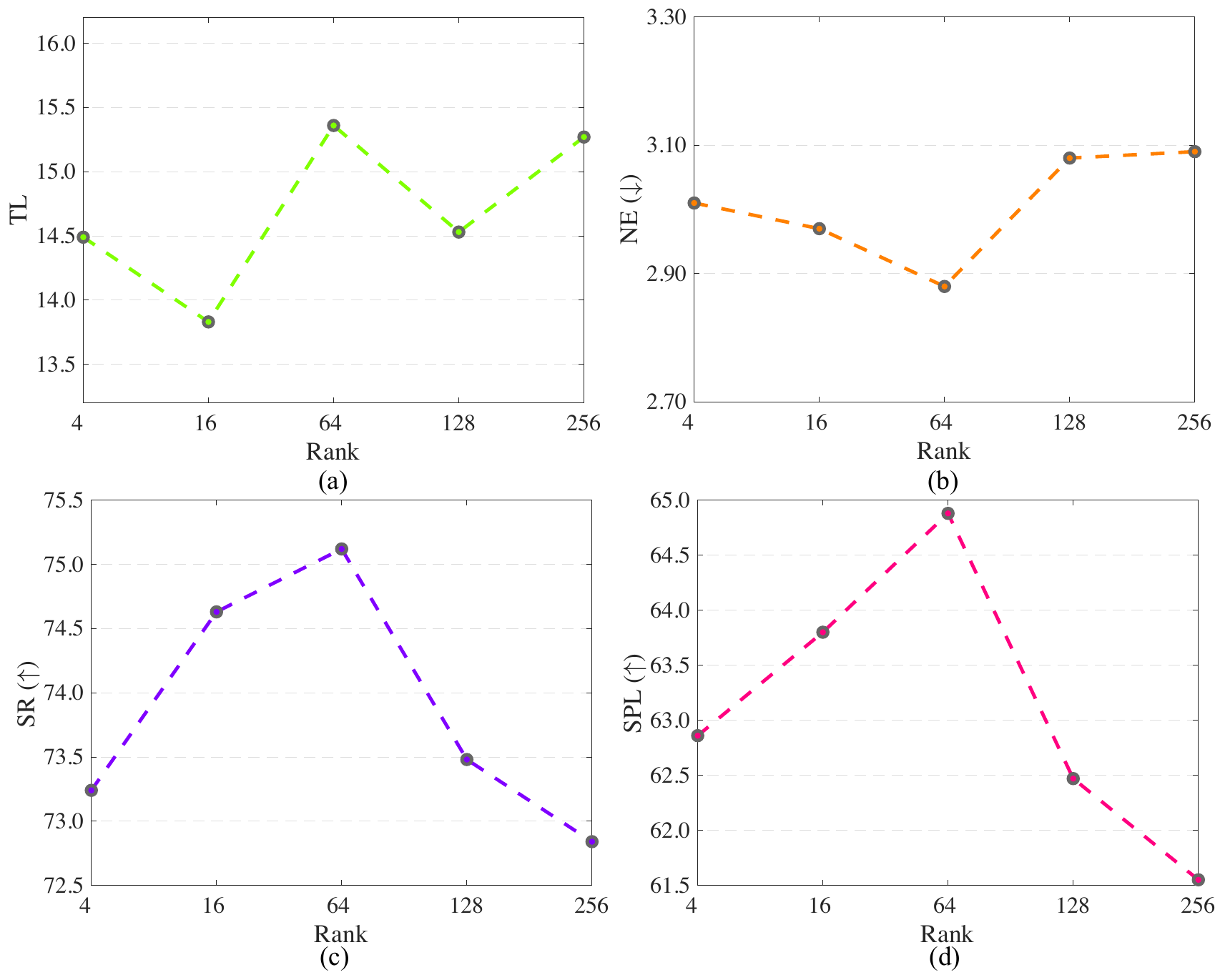}
  \caption{The experimental results for various ranks $r$ of domain adaption module in terms of TL, NE, SR, and SPL.}
  \label{fig:different_ranks}

\end{figure}

\textbf{Impact of Rank of Domain Adaptation Module of PanoGen++.}
To examine the effects of varying the rank used in training, we analyzed the results from VLN agents with different ranks $r$.
This module facilitates the training of certain dense layers in a neural network by optimizing rank decomposition matrices of the dense layers' changes during adaptation, while keeping the pre-trained weights of the diffusion model frozen.
As shown in Figure~\ref{fig:different_ranks}, as the rank increases to 64, the agent achieves the best performance. 
This suggests that the low-rank adaptation matrix potentially amplifies the important features for domain-specific knowledge that were emphasized in the general pre-trained model. 
Furthermore, a lower rank $r$ limits the expressive capacity of the low-rank matrices, preventing them from adequately capturing the features of the environments. 
Conversely, a higher rank $r$ could result in parameter redundancy, increasing the model's complexity without significantly enhancing its expressive capacity, potentially introducing noise and adversely affecting the model's generalization performance. 
Increasing $r$ does not cover a more meaningful subspace, which suggests that a low-rank adaptation matrix is sufficient. 

\subsection{Limitation and Discussion}
\label{sec:lim_and_dis}

In this section, we explore the current limitations of PanoGen++ and provide a detailed analysis and discussion.

\textbf{Performance in Advanced Goal-Oriented Exploration Tasks.}
To further evaluate our approach, we conducted additional experiments on the embodied referring expressions task, which had not been previously explored in PanoGen. The REVERIE dataset~\cite{qi2020reverie} presents a more challenging task that combines navigation with object identification. Unlike R2R and R4R, which provide detailed step-by-step instructions, REVERIE focuses primarily on describing the target location and object. This requires the agent to identify specific objects with minimal guidance. As shown in Table \ref{table:table_reverie}, PanoGen++ demonstrates consistent improvements over its predecessor PanoGen~\cite{li2023panogen} across multiple metrics in both validation seen and unseen environments. Specifically, in the validation unseen split, the inpainting setting achieves SR of 45.47\% and SPL of 32.03\%, surpassing PanoGen by 2.13\% and 0.29\% respectively. The outpainting setting exhibits comparable performance with SR of 45.13\% and SPL of 30.39\%.
Despite these improvements, our method still underperforms state-of-the-art approaches like AZHP~\cite{gao2023adaptive} and DUET~\cite{chen2022think}. This gap stems partly from AZHP's hierarchical planning and dividing navigation into zone selection and subgoal execution phases and from our environment generation strategy, which replaces trajectory observations at random. Such replacements may disrupt the visual-semantic continuity crucial for object-centric navigation. While our generated environments remain semantically coherent, they sometimes lose the fine-grained visual cues needed for accurate object recognition and localization. These challenges are especially problematic in REVERIE, which relies on abstract, goal-oriented instructions rather than R2R's detailed guidance. Success in REVERIE demands robust navigation, object recognition, and spatial reasoning. Although our approach enhances visual diversity, it struggles to preserve the precise semantics required for complex instruction following.

\begin{figure}[t]
  \centering
\includegraphics[width=0.42\textwidth]{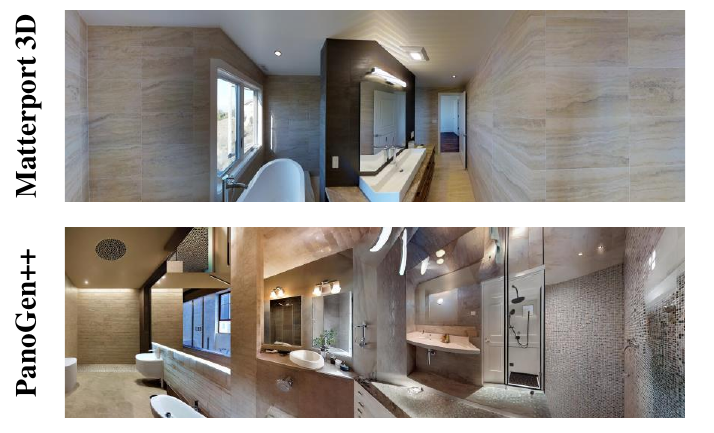}
  \caption{Comparison of a Matterport3D reference panorama and a bathroom panorama generated by PanoGen++.}
  \label{fig:bad_case}

\end{figure}

\textbf{Failure Case Analysis.} To better understand the limitations of PanoGen++, we conducted a detailed failure case analysis. Figure~\ref{fig:bad_case} illustrates a generated bathroom compared to its Matterport3D reference. The comparison highlights several key issues: geometric inconsistencies, particularly along panorama edges, and noticeable quality degradation in specific regions, such as blurriness in the rightmost section. Additionally, the generated wall textures lack the uniformity present in the reference image, potentially impacting navigation task performance. These shortcomings stem from inherent challenges in maintaining geometric consistency within 360-degree images and adapting the pre-trained model to VLN tasks. Furthermore, BLIP-2 occasionally misidentifies structural elements, such as walls, pointing to areas that require further refinement.

\section{Conclusion}
\label{sec:Conclusion}
This paper presents PanoGen++, a novel framework that combines the advantages of pre-trained generation models with domain-specific fine-tuning to generate panoramic environments tailored for VLN tasks. Our framework, incorporating masked image inpainting and recursive image outpainting, effectively produces diverse environments featuring varied room layouts, structures, and objects. These synthetic panoramic environments are utilized during the pre-training and fine-tuning stages of VLN tasks. Empirical results indicate that PanoGen++ sets new state-of-the-art benchmarks on the R2R, R4R, and CVDN datasets, underscoring its efficacy in enhancing agents' generalization to previously unseen environments. Despite PanoGen++'s effective integration of domain-specific knowledge, the diversity of the generated environments is still limited by the quality and range of the image-text pairs employed in training. Moreover, integrating multiple domain adaption modules could significantly augment our approach's capabilities. Future research could focus on incorporating more diverse datasets to further enhance environmental variability. 
Future research may explore integrating more diverse datasets and advanced text-to-image generation models to improve further environmental variability and the overall quality of VLN-specific environment generation.

\bibliographystyle{IEEEtranN}

\bibliography{PanoGen++-refs}

\end{document}